%% file: main.tex
\documentclass[10pt,twocolumn,letterpaper]{article}

\usepackage[pagenumbers]{cvpr} 

\input{preamble}

\definecolor{cvprblue}{rgb}{0.21,0.49,0.74}
\usepackage[pagebackref,breaklinks,colorlinks,allcolors=cvprblue]{hyperref}

\def\confName{CVPR}
\def\confYear{2027}

\title{Social Interaction Video Generation Needs a Social Plan
\vspace{-11pt}}

\author{Hamza Tahboub \quad Hanhui Wang \quad Huaizu Jiang\\
Northeastern University\\
{\tt\small \{tahboub.h, wang.hanh, h.jiang\}@northeastern.edu}
\vspace{-9pt}
}

\begin{document}
\twocolumn[{%
\renewcommand\twocolumn[1][]{#1}%
\maketitle
\vspace{-1.2em}
\begin{center}
  \includegraphics[width=\textwidth]{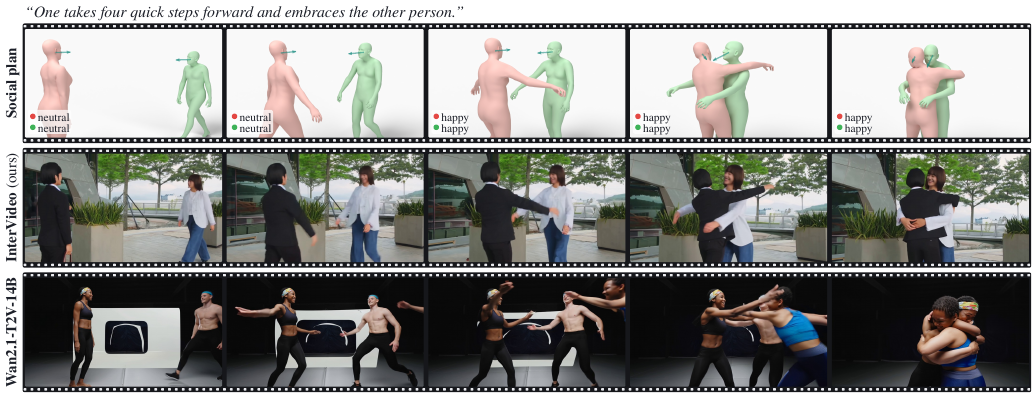}
  \captionof{figure}{\textbf{\method plans the social behavior of every person before rendering the video.}
  The social plan (top) holds each person's 3D motion, gaze (arrows), and emotion, and our video (middle) follows it.
  Given the same caption and seed but no plan, the video model our renderer is built on (bottom) adds a third person, who merges into the embrace.}
  \label{fig:teaser}
\end{center}%
\vspace{0.6em}
}]
\input{sec/0_abstract}    
\input{sec/1_intro}
\input{sec/2_related}
\input{sec/3_methods}
\input{sec/4_experiments}
\input{sec/5_conclusion}

{
    \small
    \bibliographystyle{ieeenat_fullname}
    \bibliography{main}
}

\input{sec/X_suppl}

\end{document}

%% file: preamble.tex
\newcommand{\pmv}[1]{{\scriptsize$\,\pm$#1}}
\newcommand{\method}{InterVideo\xspace}
\newcommand{\testset}{InterTest\xspace}
\newcommand{\testn}{400\xspace}

\newcommand{\affectsmall}{EmoTest\xspace}

\makeatletter
\renewcommand\paragraph{\@startsection{paragraph}{4}{\z@}{4pt \@plus 2pt \@minus 1pt}{-1em}{\normalfont\normalsize\bfseries}}
\makeatother

\usepackage{lipsum}
\usepackage{booktabs}
\usepackage{multirow}
\usepackage{pifont}
\newcommand{\cmark}{\ding{51}}
\newcommand{\xmark}{\ding{55}}
\newcommand{\gr}[1]{\textcolor{gray}{#1}}

%% file: sec/0_abstract.tex
\begin{abstract}
We address text-driven social interaction video generation, where a caption names only the action and social norms decide the rest, such as who responds and when, where each person looks, and how each person feels.
Although video generation has advanced significantly in the past few years, existing models still often get these interactions wrong, adding or dropping people, leaving actions unanswered, and performing actions out of order.
We present \method, which represents social behavior as a \emph{social plan} of every person's 3D motion, gaze, and emotion.
A vision-language model (VLM) trained on motion capture and in-the-wild videos generates the plan's motion and gaze from the caption, a learned critic chooses among its samples, and a larger VLM infers emotion.
A video diffusion model then generates the video from the plan, following each person’s pose, gaze, and emotion.
On \testn held-out two-person captions, \mbox{\method} depicts the interactions more faithfully than the video model it is built on (VQA accuracy .709 \vs{} .644) at comparable visual quality, with the largest gains in the people's actions, interactions, and gaze.
\end{abstract}

%% file: sec/1_intro.tex
\vspace{-7pt}
\section{Introduction}
\label{sec:intro}

We study text-driven social-interaction video generation. Beyond the laws of physics that physical interactions follow, social interactions are also bound by social norms and common sense. Physics keeps two people who embrace from passing through each other, but norms decide that the other person returns the embrace, that they look at each other as they approach, and that both look happy when they meet (\cref{fig:teaser}). A caption usually names only the action, and a model must infer the rest for every person from these norms, including who responds and when, where each person looks, and how each person feels. Despite significant advances \cite{liang2025video,lingbot-world,sun2026worldplay,wan2025}, existing video models still fall short on social interactions, adding or dropping people, leaving actions unanswered, and performing actions out of order (\cref{fig:teaser}, bottom row) \cite{ouyang2026socialdirector,peng2026svbench}.

These models depend on captions both during training and at inference. During training, each clip comes with a caption but no labels for individual people, so the model must learn social norms from the pixels and a description of the whole clip. In full-body scenes, gaze and expression occupy few of those pixels, with faces covering a small portion of each frame in our training clips, and the denoising loss, averaged over all positions of the latent video \cite{wan2025}, gives them a correspondingly small share of the training signal. Learning norms this way is data-inefficient, and even models trained on billions of images and videos \cite{wan2025} often violate them. At inference, every video token attends to the same caption through cross-attention \cite{wan2025}, and the caption gives no person's position, timing, gaze, or emotion. The model must fill in all of these from norms it has learned inefficiently.

To overcome these limitations, prior work generates video from signals richer than a caption. On the one hand, controllable generation produces videos of people from signals given for each person, in the form of poses \cite{chen2026dancetogether,ling2025everybodydance,hu2026motionweaver}, speech \cite{kong2025let,wang2025interacthuman,agrawal2025seamless}, or a timed script of actions \cite{ouyang2026socialdirector}, and general frameworks accept several kinds of control \cite{jiang2025vace}. These methods show that video models can follow such signals closely, but the signals already contain how each person responds to the others. Deciding this behavior from a caption is the central part of the problem because it requires applying social norms to each person's responses. Other methods derive richer signals from the caption itself. Methods that generate body motion for video focus on single-person generation \cite{Song_2025_BMVC,wang2025humandreamer,nam2025cameo,wang2026mosa}, where motion does not have to take social norms into account. Methods that plan with a language model generate layouts \cite{lin2023videodirectorgpt}, keyframe tokens \cite{huang2025plan}, or text schedules \cite{jiang2026omnihuman15}, which leave the video model to decide how each person moves and responds, returning to the same weak supervision. Social behavior in video, meaning how each person acts toward and responds to others, is thus supplied as input, left out, or left to the video model and its weak supervision. Learning it for video generation remains an open problem.
 
To bridge this gap, we propose \method, which represents social behavior as a social plan of every person's 3D motion, gaze, and emotion. A frozen video diffusion model then generates the video from the plan, following each person's pose through its control branch \cite{jiang2025vace} and their gaze and emotion through new adapters trained with per-person labels. Written by a \emph{social planner} from the caption before any frame exists, the plan fixes who takes part and how they act toward and respond to one another. The social planner comprises a motion and gaze generator and an emotion director. We instantiate the generator with a vision-language model (VLM) \cite{bai2025qwen3vl}, whose pretraining brings spatial priors as well as knowledge of social norms \cite{yuan2024measuring}. We fine-tune it to generate the plan as discrete tokens \cite{guo2024momask,zhang2023t2m}, one person at a time \cite{yu2026socialgen}, letting each person respond to those already generated. Planning social behavior separately from the video lets the generator learn from data that label each person's behavior directly. We use motion capture for 3D motion of interactions, and we reconstruct people from in-the-wild videos in 3D to add motion and gaze. We plan emotion with our emotion director, a larger VLM with stronger understanding of emotions, which infers each person's emotion from the caption and planned motion. At inference, our plans give the video model each person's position, timing, gaze, and emotion, which it would otherwise have to infer from the caption alone.

Because a caption leaves much of the behavior unstated, the planner can produce many plausible social plans for the same caption. To choose the one that carries out the caption most completely, we train a critic that reads each person's planned motion together with the caption and scores how well they match. The critic encodes the caption with a language model and learns to tell real interactions from close alternatives, including the planner's own samples, altered copies of real interactions, and motion from similar captions. When more test-time computation is available, \method samples many plans, keeps those that are physically valid, and generates the video from the one ranked highest by a learned critic. The critic thus lets \method use additional test-time computation to obtain social plans that follow the caption more closely, without changing the planner or the video model.

We evaluate on \testn held-out two-person captions by asking a frontier multimodal model questions about each caption and measuring how many it answers correctly from the video \cite{cho2024davidsonian,hu2023tifa}. Videos generated from a single social plan already yield more correct answers than those of all four video models we compare against, including the backbone (.682 \vs{} .644), and with the critic, the accuracy rises to .709. The gains are largest in the number of people, actions, interactions, and gaze, which the social plan fixes before any frame is generated. Ablating the motion, gaze, or emotion streams lowers the accuracy on the relevant questions.
 
To summarize, our contributions are as follows.
\begin{itemize}[leftmargin=10pt,labelsep=5pt,itemsep=1pt,topsep=2pt,parsep=0pt]
\item We introduce a structured social plan that explicitly represents per-person 3D motion, gaze, and emotion, enabling coordinated social behavior to be planned separately from video generation.
\item We propose a learned critic that evaluates the alignment between planned interactions and text, enabling test-time selection among candidate plans before video generation.
\item We demonstrate that \method improves prompt following over strong video-generation baselines on two-person social interactions, with ablations showing the contributions of its individual planning and control components.
\end{itemize}

%% file: sec/2_related.tex
\section{Related Work}
\label{sec:related}

\textbf{Text-to-motion generation.}
Text-to-motion models generate human motion with diffusion \citep{tevet2022human} or as discrete tokens from a vector-quantized tokenizer \citep{zhang2023t2m, guo2024momask}. MotionGPT \citep{jiang2023motiongpt} trains a language model to read and write such tokens like text. For two people, InterGen \citep{liang2024intergen} and InterMask \citep{javed2024intermask} learn from the motion capture of InterHuman and Inter-X \citep{xu2023inter}, and SocialGen \citep{yu2026socialgen} extends MotionGPT to any number of people. See \citep{sui2026survey} for a more detailed survey. None of these models represents gaze or emotion, chooses among its samples, or renders video. MotionCritic \citep{wang2025aligning} learns the perceived quality of motion from human preferences, but uses it to fine-tune a generator rather than to choose among its samples. Our planner uses discrete tokens and SocialGen's person-by-person layout, and is the first to also learn from multi-person motion lifted from in-the-wild video.

\textbf{Gaze and emotion.}
Gaze shows who each person attends to \citep{frischen2007gaze} and facial expression how they feel. Generative models that control them tend to do so for single faces driven by speech, such as eye movement synthesized from audio \citep{zhuang2026talkingeyes}, continuous valence and arousal for a talking head \citep{jia2026cetalk}, or both \citep{liang2026emotional}. STEER \citep{teotia2026steer} is the closest to our choice of signals, factorizing conversational behavior into gaze, head rhythm, and emotion, but for one audio-driven head avatar. \method plans gaze, valence, and arousal \citep{russell1980circumplex} for every person in a full-body scene from text alone.

\textbf{Human video generation.}
Video diffusion transformers \citep{wan2025} accept control signals through adapters and unified conditioning branches such as VACE \citep{jiang2025vace}, which our renderer builds on. Pose-driven animation \citep{li2024animate, ma2025controllable} makes a person follow a pose sequence that is usually extracted from an existing video, so what the person does is given rather than generated. Multi-person methods bind identities and audio to the right person \citep{wang2025interacthuman}, and audio-driven models produce full-body \citep{lin2025omnihuman} and two-person \citep{pan2026interdyad} video. SocialDirector \citep{ouyang2026socialdirector} shares our goal of controlling who acts, when, and toward whom, but does so at inference by masking cross-attention to parts of the prompt, without modeling motion, gaze, or emotion. To our knowledge, no video generator conditions on each person's gaze, valence, and arousal in multi-person scenes.

\begin{figure*}[t]
  \centering
  \includegraphics[width=\textwidth]{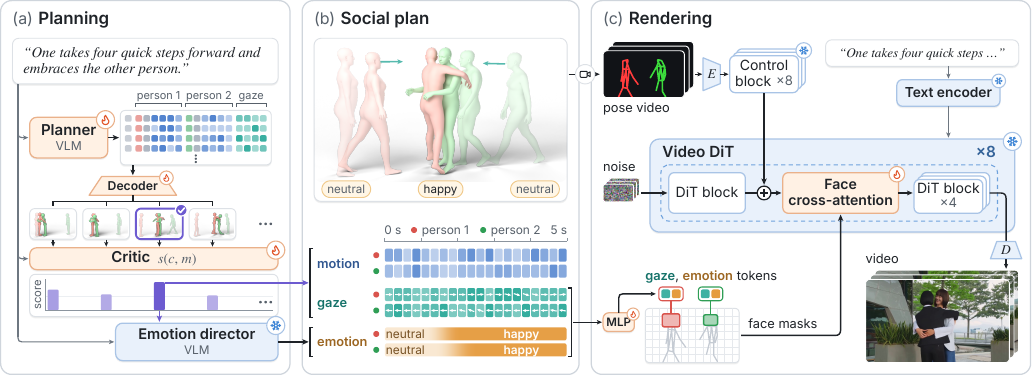}
  \caption{\textbf{Overview of \method.}
    (a)~A fine-tuned VLM samples many plans, a critic picks the one that best fits the caption, and a frozen VLM adds each person's emotion.
    (b)~The social plan holds every person's 3D motion, gaze (arrows), and emotion.
    (c)~A frozen video model renders the plan from a pose video, and new cross-attention layers let each face attend to its person's gaze and emotion.
    The plan is complete before any frame is generated, and the video model stays frozen.
    }
  \label{fig:overview}
\end{figure*}

\textbf{Planning before rendering.}
Several systems let a language model write a plan that conditions the video generator, such as a scene layout \citep{lin2023videodirectorgpt} or camera keyframes in a readable language \citep{kizil2026auteur}. Closest to ours, DirectorLLM \citep{Song_2025_BMVC} plans boxes and image-plane keypoints for the single largest person in the frame, and HumanDreamer \citep{wang2025humandreamer} generates one person's pose sequence from text before rendering it. Both plan only the 2D pose of a single person. To our knowledge, no prior method plans even the 3D motion of several people and renders a video from the plan (\cref{tab:position}), whereas \method plans their 3D motion, gaze, and emotion together. Other systems plan in a learned latent space \citep{huang2025plan} or in semantic tokens that a frozen generator reads through cross-attention \citep{tu2026baton}, or generate motion and video jointly \citep{zhao2026comovi}. Latent plans need no labels, but they cannot be read, edited, or checked against off-the-shelf estimators before rendering. Planning first also makes it cheap to choose among samples. Test-time scaling, which spends inference compute on search \citep{snell2025scaling}, has been applied to image \citep{ma2025scaling} and video generation \citep{liu2025videot1, oshima2025inference, he2025scaling} by searching over initial noises or denoising paths for the sample that a verifier, such as a video reward model \citep{liu2025improving}, scores highest. These methods search over videos, costing denoising steps of a large video model for every candidate. We search over plans instead. Our 3D plans of several people take a fraction of a second to sample and are compared by a learned critic before anything is rendered. SketchVerify \citep{huang2025sketchverify} also searches over plans, but over 2D trajectories of objects, and it still renders each one as a rough video sketch for a vision-language model to judge.

\begin{table}[t]
\centering\footnotesize
\setlength{\tabcolsep}{0pt}
\caption{\textbf{Closest methods.}
Only \method plans the motion, gaze, and emotion of several people and renders them as video.}
\begin{tabular*}{\columnwidth}{@{\extracolsep{\fill}}lccccc@{}}
\toprule
 & & & & Body & Renders \\
 & Interaction & Emotion & Gaze & motion & video \\
\midrule
InterGen \citep{liang2024intergen} & \cmark & \xmark & \xmark & \cmark & \xmark \\
InterMask \citep{javed2024intermask} & \cmark & \xmark & \xmark & \cmark & \xmark \\
SocialGen \citep{yu2026socialgen} & \cmark & \xmark & \xmark & \cmark & \xmark \\
DirectorLLM \citep{Song_2025_BMVC} & \xmark & \xmark & \xmark & \cmark & \cmark \\
HumanDreamer \citep{wang2025humandreamer} & \xmark & \xmark & \xmark & \cmark & \cmark \\
SocialDirector \citep{ouyang2026socialdirector} & \cmark & \xmark & \xmark & \xmark & \cmark \\
STEER \citep{teotia2026steer} & \xmark & \cmark & \cmark & \xmark & \cmark \\
\midrule
\method & \cmark & \cmark & \cmark & \cmark & \cmark \\
\bottomrule
\end{tabular*}

\label{tab:position}
\end{table}

%% file: sec/3_methods.tex
\section{Methods}
\label{sec:method}

Given a caption $c$, \method generates a five-second video ($T{=}81$ frames at 16\,fps) by writing a social plan and then rendering it (\cref{fig:overview}).
A generator writes how every person moves and where they look, an emotion director adds what each face expresses, and a renderer produces a video that follows the completed plan.
Optionally, a learned critic chooses among many sampled plans before the emotion is added.

\subsection{Social Plans}
\label{sec:formulation}

A plan for $K$ people is $P=\{(m_k,g_k,e_k)\}_{k=1}^{K}$, with three streams per person.
The motion $m_k\in\mathbb{R}^{T\times 18\times 3}$ holds the 3D positions of the 18 OpenPose body joints \citep{cao2017realtime} in a coordinate frame shared by all people.
The gaze $g_k\in\mathbb{R}^{T\times 3}$ holds, for each frame, the unit vector along which person $k$ looks, in the same coordinate frame as the motion.
Finally, the emotion $e_k\in[-1,1]^{T\times 2}$ holds a valence (how positive) and an arousal (how intense) per frame \citep{russell1980circumplex}, which, unlike rigid emotion categories, can change gradually over a clip.
All three streams can be measured on real video by off-the-shelf estimators, so neither the planner nor the renderer needs hand-annotated plans (\cref{sec:data}).

Our social plan is in 3D and does not depend on a pre-defined camera.
It states depth and contact directly, has no occlusion, can be filmed from any viewpoint after planning, and holds gaze and emotion even for people whose faces the camera does not see.
In contrast, a planner of 2D keypoints \citep{Song_2025_BMVC} must learn the camera of every training video along with the motion, and sees depth, contact, and occlusion only through a projection.
Experimental results show that our 3D social planner matches or exceeds a 2D planner trained with the same recipe (\cref{sec:supp_design}). 

\begin{figure*}[t]
\centering
\includegraphics[width=\textwidth]{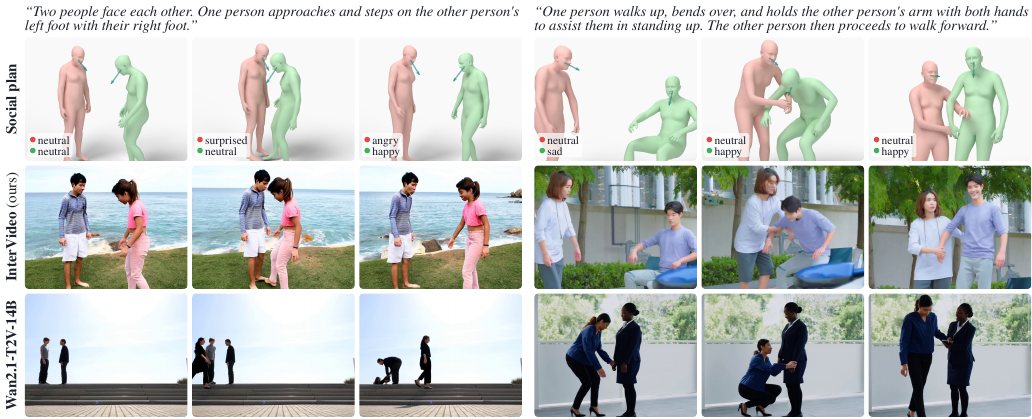}
\caption{\textbf{Qualitative results.}
Our plans have one person step on the foot of the other, who turns from surprised to angry, and have one person help a seated man stand up as he turns from sad to happy.
Given the same captions and seeds, the backbone adds a third person in the first scene and seats nobody in the second, so neither interaction takes place.}
\label{fig:qual}
\end{figure*}

\subsection{Social Planner}
\label{sec:planner}
The social planner consists of two VLMs. A fine-tuned motion and gaze generator synthesizes each person's motion and gaze, and a larger, frozen emotion director infers their emotion.

\noindent\textbf{Motion and gaze generator.}
We fully fine-tune Qwen3-VL-4B-Instruct \citep{bai2025qwen3vl} to generate discrete plan tokens (\cref{fig:overview}a).
Each person's motion is represented by motion tokens for local body movements and placement tokens for relative positions and orientations.
We normalize motion by the hips and facing direction, and use a T2M-GPT-style convolutional encoder \citep{zhang2023t2m} to encode every four frames into three residual body codes \citep{guo2024momask} and one head code for finer head motion.
On held-out motion capture, the head's motion relative to the neck is reconstructed with an error of 33\,mm by one body codebook and 27\,mm by two, against 14\,mm by one body and one head codebook. The head is critical for representing social motion, so we adopt this structure.
Placement tokens specify each person's position and heading relative to the first person, initially and every second thereafter.
Gaze is encoded as one finite scalar quantization code \citep{mentzer2024finite} per four frames, representing azimuth and elevation relative to the facing direction.

The generator autoregressively produces
\begin{equation}
  x=\big[K,\,t_1,\,t_2,\,x^{m}_{1},\dots,x^{m}_{K},\,x^{g}_{1},\dots,x^{g}_{K}\big],
  \label{eq:plan}
\end{equation}
where $K$ is the number of people, $t_1$ and $t_2$ encode the clip duration and starting position, and $x^m_k$ and $x^g_k$ are person $k$'s motion (including placement) and gaze codes.
The person count is determined first and enforced during decoding.
Following SocialGen \citep{yu2026socialgen}, each person's motion is conditioned on previously generated people, while gaze is generated last to account for everyone's positions.
The timing tokens support training on random five-second crops, and repeated placement, head codes, gaze, timing, and person count extend SocialGen's layout. 

Following MotionGPT \citep{jiang2023motiongpt}, we train the generator on caption-to-plan, plan-to-caption, and partial-plan completion tasks, using studio motion capture and motion and gaze reconstructed from in-the-wild videos (\cref{sec:data}).
A text-only model of the same size and family performs slightly worse (\cref{sec:supp_design}).
At inference, we sample with temperature 1 and classifier-free guidance \citep{ho2022classifier} of weight 3.

\noindent\textbf{Emotion director.}
\label{sec:director}
Since captions rarely specify emotions, we use a frozen Qwen3.8-27B \citep{qwen38} to infer them from the caption and planned motion (\cref{sec:affect_who}).
The model receives the caption and eight frames of planned motion rendered as labeled stick figures, and predicts time segments labeled with one of seven expressions for each person.
These are converted to valence and arousal with smooth transitions to obtain $e_k$ (\cref{sec:supp_affect}).

\subsection{Plan Critic}
\label{sec:critic}

One sampled plan may leave out part of its caption (\cref{fig:supp_critic}), and so may one sampled video.
But a video takes minutes to render, whereas sampling and scoring a plan takes a fraction of a second.
An explicit plan therefore allows a search that is impractical over videos. We sample 56 plans per caption and render only the one that a learned critic ranks best.
This optional step sits between the generator and the emotion director, which then completes the chosen plan, and improves both the plans and the videos (\cref{sec:plannereval,sec:main}).

\paragraph{Model.}
The critic $s(c,m)$ scores how well a plan's motion $m$ fits the caption $c$.
It reads only the motion, every joint's 3D position and velocity, since its training data has no gaze.
To tell who does what and in which order, its four transformer layers cross-attend to the caption's hidden states in a frozen Qwen3-1.7B \citep{yang2025qwen3}.

\paragraph{Training and selection.}
The candidates are all sampled for one caption, so the critic is trained on InterHuman and Inter-X with the InfoNCE loss \citep{oord2018representation} against near misses: the planner's own samples, altered and corrupted copies of the real motion, and the real motion of similar captions. We discard candidates whose bodies pass through each other and rank the rest by their summed ranks under an ensemble of three critics differing mainly in negatives (\cref{sec:supp_critic}).

\begin{table*}[t]
\centering\small
\setlength{\tabcolsep}{3.25pt}
\caption{\textbf{Prompt following and visual quality on \testset.}
The last row scores the critic's choice instead of one sampled plan. Our method follows the captions better than all four video models, especially in the actions, interactions, and gaze, and falls below its backbone without the planned poses. Best in bold, second best underlined.}
\begin{tabular}{@{}lcccccccccc@{}}
\toprule
 & \multicolumn{8}{c}{VQA Accuracy $\uparrow$} & \multicolumn{2}{c}{VBench $\uparrow$} \\
\cmidrule(lr){2-9}\cmidrule(l){10-11}
 & All & Count & Action & Interaction & Order & Expression & Wrong expr. & Gaze & Imaging & Aesthetic \\
\midrule
Wan2.1-T2V-14B & .644 & .825 & .598 & \underline{.587} & .339 & \textbf{.721} & .923 & .323 & .673 & \underline{.579} \\
HunyuanVideo 1.5 (8B) \citep{hunyuanvideo15} & .550 & .631 & .525 & .496 & .275 & .625 & .941 & .290 & .708 & .495 \\
CogVideoX-5B \citep{yang2024cogvideox} & .500 & .591 & .487 & .464 & .185 & .470 & \textbf{.957} & .195 & .585 & .474 \\
LTX-2.5 (22B) \citep{hacohen2026ltx} & .549 & .399 & .548 & .533 & \textbf{.405} & .652 & .912 & \underline{.488} & .711 & \textbf{.592} \\
\midrule
\method without pose & .540 & .731 & .487 & .541 & .243 & .530 & .925 & .326 & .705 & .548 \\
\method & \underline{.682} & \underline{.925} & \underline{.652} & .585 & .356 & \underline{.681} & .925 & \textbf{.491} & \textbf{.739} & .537 \\
\method + critic & \textbf{.709} & \textbf{.945} & \textbf{.684} & \textbf{.659} & \underline{.404} & .679 & \underline{.948} & \underline{.488} & \underline{.736} & .536 \\
\bottomrule
\end{tabular}

\label{tab:main}
\end{table*}

\subsection{Social Video Renderer}
\label{sec:renderer}

To preserve a pre-trained video model's visual quality, the renderer keeps it frozen. We use Wan2.1-VACE-14B \citep{jiang2025vace,wan2025}, which adds a control branch to Wan2.1-T2V-14B without changing its weights, and train 168M new cross-attention injection parameters.

\paragraph{Pose.}
The control branch was trained to follow pose videos, so a virtual camera films the planned motion, with each person's skeleton drawn in its own color (\cref{sec:supp_impl}).

\paragraph{Gaze and emotion.}
A pose video shows neither gaze nor emotion, and both must reach the right person's face, so we add a cross-attention adapter after each of the eight DiT blocks that receive the control signal (\cref{fig:overview}c).
For each person $k$ and latent frame $f$, which covers four video frames, the planned gaze, rotated into camera coordinates, and the planned emotion are each averaged and embedded into one token, $g_{k,f}$ and $e_{k,f}$.
Let $\mathcal{F}_{k,f}$ be the video tokens of latent frame $f$ inside person $k$'s face region, a box around the head joints of the projected plan.
An adapter updates the hidden state $h_i$ of each $i\in\mathcal{F}_{k,f}$ as
\begin{equation}
  h_i \leftarrow h_i + W_o\,\mathrm{Attn}\big(h_i,\,[g_{k,f},\,e_{k,f}],\,[g_{k,f},\,e_{k,f}]\big),
  \label{eq:adapter}
\end{equation}
with $h_i$ as the query, the person's two tokens as the keys and values, and a zero-initialized output projection $W_o$.
All other video tokens are left unchanged.
Each face thus attends only to its own person's tokens of the same latent frame, which binds each stream to the right person without a person embedding.

\paragraph{Training.}
The adapters are trained with flow matching \citep{lipman2023flow} on 86,410 in-the-wild clips and the poses, gaze, and emotion read from them (\cref{sec:data}).
Gaze, emotion, and the caption are each dropped on part of the samples, so that each can be given its own guidance weight (\cref{sec:supp_impl}).

\paragraph{Guidance.}
With $v_\varnothing$ (the prediction with the pose video but without gaze, emotion, or caption), $v_{g}$ (with gaze added), $v_{ge}$ (with gaze and emotion), and $v$ (with everything), the sampler uses
\begin{equation}
  \hat v = v_\varnothing + s_g(v_{g}-v_\varnothing) + s_e(v_{ge}-v_{g}) + s_t(v-v_{ge}),
  \label{eq:guidance}
\end{equation}
with $s_g{=}s_e{=}8$ and text guidance $s_t{=}5$, the last rescaled as in \citet{lin2024common} with factor 0.7.
Each difference isolates what one condition adds to those before it, so each stream's strength can be set on its own.

%% file: sec/4_experiments.tex
\section{Experiments}
\label{sec:experiments}

We evaluate \method in three ways: its videos against text-to-video models (\cref{sec:main}), its plans against two-person motion generators (\cref{sec:plannereval}), and each stream of the plan by leaving it out (\cref{sec:ablation}).

\subsection{Setup}
\label{sec:eval}

\paragraph{Training data.}
\label{sec:data}
The planner is trained on motion capture from InterHuman \citep{liang2024intergen}, Inter-X \citep{xu2023inter}, HOI-M$^3$ \citep{zhang2024hoi}, and CMU Panoptic \citep{joo2019social}, and on 120,109 five-second windows of OpenHumanVid \citep{li2024openhumanvid} lifted to 3D with SAM~3 \citep{carion2025sam3} and SAM 3D Body \citep{sam3dbody2026}. The official test splits of InterHuman and Inter-X are held out. The renderer is trained on 86,410 OpenHumanVid clips labeled frame by frame with tracked 2D keypoints \citep{jocher2024yolo11,jiang2023rtmpose}, gaze \citep{vuillecard2025enhancing}, and valence and arousal \citep{savchenko2022classifying}. Details of the data, training, and sampling are in \cref{sec:supp_data,sec:supp_impl}.

\paragraph{Test captions.}
\testset has \testn two-person captions from the official test splits, half from InterHuman and half from Inter-X, none of which occurs in the training data. Most imply only a calm or focused face, so we test emotion on \affectsmall, 40 two-person captions that we curated. Half imply each person's emotion without naming it, and in the other half, an expression changes during the clip.

\paragraph{Baselines.}
Our main baseline is Wan2.1-T2V-14B \citep{wan2025}, the backbone of our renderer. We also compare with HunyuanVideo~1.5 \citep{hunyuanvideo15}, CogVideoX-5B \citep{yang2024cogvideox}, and LTX-2.5 \citep{hacohen2026ltx}. We compare pure motion with the two-person generators InterGen \citep{liang2024intergen} and InterMask \citep{javed2024intermask}.
SocialGen's code has not been released at the time of writing.

\paragraph{Metrics.}
We measure how well a video follows its caption by visual question answering \citep{hu2023tifa,cho2024davidsonian}.
For each caption, a judge, GPT-6.1 Sol \citep{openai2026gpt61sol}, writes yes/no questions about the number of people, each person's action, and, when the caption states them, the interaction (contact and direction) and the order of events. More details are in \cref{sec:supp_eval}.
Visual quality is measured with the imaging and aesthetic quality scores of VBench \citep{huang2023vbench}. Motion alone is evaluated on the full InterHuman and Inter-X test splits with the text-to-motion protocol of \citet{guo2022generating} averaged over 20 generations per caption. The judge also scores the plans as stick-figure videos on 2,686 held-out captions.

\begin{table*}[t]
\centering\footnotesize
\setlength{\tabcolsep}{4.2pt}
\caption{\textbf{Social text-to-motion generation} on the InterHuman and Inter-X test sets. Evaluators are trained in our joint representation, so values differ from published ones. $\pm$: 95\% confidence interval over 20 generations. VQA Acc.: the judge's accuracy on the motion shown as stick-figure videos. One sampled plan is close to the strongest generators with a lower FID, and the critic's choice leads on both datasets.}
\begin{tabular}{@{}llccccccc@{}}
\toprule
 & & \multicolumn{3}{c}{R-Precision $\uparrow$} & & & & \\
\cmidrule(lr){3-5}
Dataset & Method & Top-1 & Top-2 & Top-3 & FID $\downarrow$ & MM-Dist $\downarrow$ & Diversity $\rightarrow$ & VQA Acc. $\uparrow$ \\
\midrule
\multirow{6}{*}[-4.3pt]{\rotatebox[origin=c]{90}{InterHuman}}
 & \gr{Ground truth} & \gr{.336\pmv{.005}} & \gr{.502\pmv{.005}} & \gr{.602\pmv{.004}} & \gr{0.01\pmv{.00}} & \gr{5.70\pmv{.02}} & \gr{13.76\pmv{.16}} & \gr{--} \\
 & \gr{Our tokenizer (recon.)} & \gr{.330\pmv{.005}} & \gr{.491\pmv{.006}} & \gr{.587\pmv{.005}} & \gr{0.10\pmv{.01}} & \gr{5.79\pmv{.02}} & \gr{13.55\pmv{.15}} & \gr{--} \\
\cmidrule(l){2-9}
 & InterGen & .320\pmv{.005} & .473\pmv{.006} & .572\pmv{.005} & 3.83\pmv{.09} & 6.01\pmv{.03} & 12.76\pmv{.14} & .623 \\
 & InterMask & .303\pmv{.005} & .460\pmv{.006} & .555\pmv{.005} & 2.83\pmv{.13} & 6.14\pmv{.03} & 13.08\pmv{.10} & .614 \\
\cmidrule(l){2-9}
 & \method planner & .306\pmv{.006} & .457\pmv{.006} & .555\pmv{.006} & 0.98\pmv{.06} & 6.01\pmv{.03} & 13.00\pmv{.10} & .692 \\
 & \method planner + critic & \textbf{.343}\pmv{.005} & \textbf{.506}\pmv{.006} & \textbf{.604}\pmv{.005} & \textbf{0.68}\pmv{.04} & \textbf{5.51}\pmv{.03} & \textbf{13.27}\pmv{.11} & \textbf{.700} \\
\midrule
\multirow{5}{*}[-4.3pt]{\rotatebox[origin=c]{90}{Inter-X}}
 & \gr{Ground truth} & \gr{.613\pmv{.005}} & \gr{.790\pmv{.004}} & \gr{.861\pmv{.004}} & \gr{0.00\pmv{.00}} & \gr{3.79\pmv{.02}} & \gr{15.63\pmv{.14}} & \gr{--} \\
 & \gr{Our tokenizer (recon.)} & \gr{.607\pmv{.004}} & \gr{.786\pmv{.005}} & \gr{.856\pmv{.004}} & \gr{0.02\pmv{.00}} & \gr{3.86\pmv{.02}} & \gr{15.67\pmv{.14}} & \gr{--} \\
\cmidrule(l){2-9}
 & InterMask & .607\pmv{.004} & .786\pmv{.004} & .860\pmv{.003} & 0.47\pmv{.02} & 3.88\pmv{.03} & \textbf{15.71}\pmv{.13} & .643 \\
\cmidrule(l){2-9}
 & \method planner & .591\pmv{.005} & .772\pmv{.004} & .848\pmv{.003} & 0.38\pmv{.02} & 3.98\pmv{.02} & 15.54\pmv{.14} & .665 \\
 & \method planner + critic & \textbf{.658}\pmv{.003} & \textbf{.839}\pmv{.003} & \textbf{.904}\pmv{.003} & \textbf{0.22}\pmv{.01} & \textbf{3.25}\pmv{.02} & 15.94\pmv{.12} & \textbf{.698} \\
\bottomrule
\end{tabular}
\label{tab:motion}
\end{table*}

\subsection{Video Results}
\label{sec:main}

\cref{tab:main} compares \method with the four video models on \testset. \method reaches the highest VQA accuracy: .682 with one sampled plan per caption, against .644 for its backbone, the strongest of the four, and .500 to .550 for the other three.
The largest gains over the backbone are in the number of people, the actions, and gaze. The plan fixes who is in the scene, and the judge finds the wrong number of people in 7.5\% of our videos, against 17.5\% from the backbone, which often adds a bystander or merges two people (\cref{fig:teaser,fig:qual}).
Accuracy on the actions rises from .598 to .652, and on gaze shifts from .323 to .491.

Rendering the critic's choice among 56 plans raises accuracy to .709. The critic adds most in the interaction (.585 to .659) and the order of events (.356 to .404), where a single sample is about level with the backbone. A sampled plan often leaves out part of the caption, and the critic prefers a plan that includes it (\cref{fig:supp_critic}).

The backbone is ahead only on expressions (.721 \vs{} .681). This may be because our fixed full shots show smaller faces than the backbone's videos, and we find that the deficit lies particularly in the videos where they are much smaller. \cref{sec:ablation} tests the emotion stream in closer shots. Among the other models, LTX-2.5 matches \method with the critic on gaze shifts (.488) and the order of events (.405), but shows the right number of people in fewer than half of its videos (.399). Visual quality is comparable: \method is higher on VBench imaging quality (.739 \vs{} .673) and lower on aesthetic quality (.537 \vs{} .579).
More examples, including failure cases, are in \cref{sec:supp_qual}.

\begin{figure}[!b]
\centering
\includegraphics[width=\columnwidth, trim=0 7pt 0 0, clip]{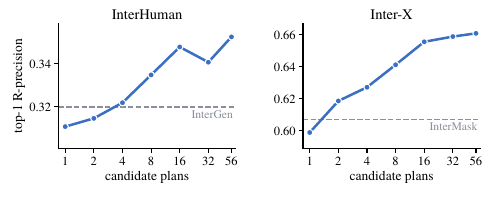}
\caption{\textbf{Choosing among plans.} Top-1 R-precision of the critic's choice against the number of sampled plans, on the InterHuman and Inter-X test sets (\cref{tab:supp_ncand}). A single plan is just below the strongest two-person generator (dashed), and a choice among eight already passes it on both datasets.}
\label{fig:critic}
\end{figure}

\subsection{Motion Results}
\label{sec:plannereval}

\cref{tab:motion} evaluates the plans as motion, before any rendering, against the two-person motion generators.

\paragraph{A single plan.}
With one sample per caption, the planner's R-precision is within .02 of the strongest baseline generator on each dataset, and its FID is the lowest. The judge, shown each motion as a stick-figure video, answers more questions correctly for our plans than for theirs (.692 \vs{} .623 and .614 on InterHuman, .665 \vs{} .643 on Inter-X), mainly on the actions, the interaction, and the order of events.

\paragraph{The critic's choice.}
\label{sec:search}
We report results for the critic's choice from 56 plans in \cref{tab:motion}. Performance continues to improve at this scale while remaining vastly cheaper than rendering (\cref{tab:supp_ncand}). The critic raises top-1 R-precision from .306 to .343 on InterHuman and from .591 to .658 on Inter-X, and puts our planner ahead of the generators in R-precision, FID, and MM-Dist on both datasets. Few plans are needed to surpass the baselines (\cref{fig:critic}): a choice among four reaches the strongest generator on InterHuman and passes it on Inter-X, and a choice among eight passes both significantly (\cref{sec:supp_design}).


\subsection{What Each Stream Adds}
\label{sec:ablation}
Each ablation renders the same sampled plans with the same seeds but leaves one stream out: a blank control video replaces the pose video and a learned null token replaces the gaze or emotion tokens in every guidance term.

\paragraph{Pose.}
Without the pose video, the renderer keeps the control branch, the adapters, the plan's gaze and emotion, and our sampler. VQA accuracy on \testset nevertheless falls from .682 to .540, below the backbone's .644, and no category improves (\cref{tab:main}).
The number of people, for example, is right in .731 of the videos instead of .925.
The gain over the backbone thus comes from the planned motion.

\begin{figure}[!t]
\centering
\includegraphics[width=\columnwidth]{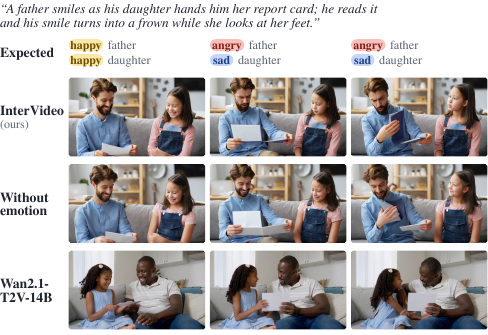}
\caption{\textbf{The emotion stream controls each person's expression.}
One \affectsmall caption rendered by \method, by \method without the emotion stream (same plan and seed), and by the backbone.
Only with the emotion stream does the father's smile turn into a frown and the daughter's fade.}
\label{fig:affect}
\end{figure}

\paragraph{Emotion.}
\label{sec:affect_who}
\cref{tab:emotion} tests the emotion stream on \affectsmall, where the camera follows the people in a medium shot. Rendered from motion and gaze alone, our videos reach an accuracy of only .567 on the implied expressions and their timing.
With the director's emotion, they reach .783, better than the backbone, and expressions that the caption rules out become as rare as in the backbone's videos. The stream sets each person's expression separately, and the expression changes when the plan says so (\cref{fig:affect}). The emotion has to be written by a larger model with better emotional capacity than our motion and gaze generator. When the generator is trained on in-the-wild labels to write the emotion itself, the videos match their captions no better than with no emotion (\cref{sec:supp_affect}).

\begin{table}[b]
\centering\footnotesize
\setlength{\tabcolsep}{5pt}
\caption{\textbf{The emotion stream.} VQA accuracy on \affectsmall.
Expression: each person shows the implied emotion at the right time. Wrong expr.: nobody shows an emotion the caption rules out.
The emotion stream raises both to or past the backbone's level.}
\begin{tabular}{@{}lcc@{}}
\toprule
 & Expression & Wrong expr. \\
\midrule
Wan2.1-T2V-14B & .758 & \textbf{.865} \\
\method without emotion & .567 & .743 \\
\method & \textbf{.783} & \textbf{.865} \\
\bottomrule
\end{tabular}

\label{tab:emotion}
\end{table}

\begin{table}[t]
\centering\footnotesize
\setlength{\tabcolsep}{1.8pt}
\caption{\textbf{Where people look.} Left: gaze statistics of the \testset videos and of 425 real two-person clips, with the mean absolute gap between them.
Right: VQA accuracy on the gaze-shift questions.
\method is the closest to real video, and without the gaze stream its gap more than doubles. Best (closest to real or highest) in bold, second best underlined.}
\begin{tabular}{@{}lccccccc@{}}
\toprule
 & \multicolumn{5}{c}{Against real video} & \multicolumn{2}{c}{Gaze shifts} \\
\cmidrule(lr){2-6}\cmidrule(l){7-8}
 & Mutual & Stare & Down & Away & Gap$\downarrow$ & Shift$\uparrow$ & Both$\uparrow$ \\
\midrule
Real videos & .204 & .160 & .318 & .178 & -- & -- & -- \\
\midrule
Wan2.1-T2V-14B & .357 & .290 & .196 & .203 & .107 & .323 & .225 \\
HunyuanVideo 1.5 & \underline{.241} & \underline{.141} & .210 & .140 & .050 & .290 & .213 \\
CogVideoX-5B & .283 & .206 & .211 & \underline{.188} & .060 & .195 & .165 \\
LTX-2.5 & \textbf{.185} & .086 & .289 & .202 & .037 & \underline{.488} & .371 \\
\midrule
\method without gaze & .314 & .209 & .258 & .225 & .066 & .425 & .326 \\
\method & .272 & .198 & \textbf{.316} & \textbf{.187} & \underline{.029} & \textbf{.491} & \textbf{.392} \\
\method + critic & .245 & \textbf{.164} & \textbf{.316} & .201 & \textbf{.017} & \underline{.488} & \underline{.374} \\
\bottomrule
\end{tabular}

\label{tab:gaze}
\end{table}

\paragraph{Gaze.}
\cref{tab:gaze} compares where people look in the \testset videos and in 425 real two-person clips with the same gaze estimator \citep{abdelrahman2023l2cs}, different from the one that labeled our training data. Mutual is the share of frames in which the two people look at each other. Stare is the share of people who look at the other for 90\% of the clip or more. Down is the share of frames with gaze more than 20$^\circ$ below horizontal. Away is the share with the head turned toward the other person but the eyes elsewhere. For all of these, closer to the real proportion is better. Gap is the mean absolute difference between a model's four statistics and those of the real clips (lower is better). Shift is the VQA accuracy on the gaze-shift questions, and Both counts a shift only when its event is also shown correctly. The backbone has people look at each other far more than real people do: in .357 of frames (real: .204), and .290 of its people stare at the other almost throughout (real: .160). Over the four measures, \method is the closest to real video (gap .029, and .017 with the critic, \vs{} .037 for LTX-2.5, the closest video model). Without the gaze stream, people look down too rarely and away too often (gap .066). With it, they do both about as often as in real clips.
The gaze-shift questions, which ask whether a person looks away from the other when an event calls for it, also show the improvement. \method scores .491, the highest of any model (LTX-2.5 .488, the others at most .323), against .425 without the gaze stream.

%% file: sec/5_conclusion.tex
\section{Conclusion and Future Work}
\label{sec:limitations}

We presented \method, which represents social behavior as a social plan of every person’s 3D motion, gaze, and emotion. A fine-tuned motion and gaze generator produces each person’s motion and gaze, a learned critic chooses among its samples, a larger emotion director adds each person’s emotion, and a frozen video model generates the video from the plan. On held-out captions, the resulting videos depicted the interactions more faithfully than the backbone’s, at comparable visual quality. They improved because a plan fixes, before any frame is generated, how many people there are, what each person does, and where each person looks. Video models render people well, and our results suggest that how those people act toward one another can be learned from data that label each person’s behavior directly, in a plan that the video model follows.

\paragraph{Future work.} Our social plan has no fingers and no objects. What the hands do and hold is left to the renderer, which does not always take it from the caption (\cref{fig:supp_fail}). 
Future work may employ SAM 3D Body \citep{sam3dbody2026}, which also recovers hands, and HOI-M$^3$ \citep{zhang2024hoi}, which records the objects people handle, to add both to the plan.
Furthermore, the generator cannot write emotion itself, and on expressions our videos do not consistently improve on the backbone's (\cref{tab:main,tab:emotion}).
In addition to native emotion, future work may consider planning the camera as well. Our work may also provide a starting point for future work in hierarchical long-form video generation.

\section*{LLM Usage Statement}

Large language models were used for assistance in experiment script implementation, debugging, visualization, literature search, and refining the manuscript. All AI-generated code, experimental outputs, and citations were verified and tested by the authors. The authors take full responsibility for the scientific accuracy, originality, and integrity of all content presented in this paper.

%% file: sec/X_suppl.tex
\clearpage
\maketitlesupplementary
\setcounter{section}{0}
\renewcommand{\thesection}{\Alph{section}}

\section{Data and Annotation}
\label{sec:supp_data}

\paragraph{Motion capture.}
The planner's interaction data come from InterHuman \citep{liang2024intergen}, Inter-X \citep{xu2023inter}, HOI-M$^3$ \citep{zhang2024hoi}, and the haggling sequences of CMU Panoptic \citep{joo2019social}.
We convert every sequence to our 18-joint skeleton with canonical bone lengths and resample it to 16\,fps.
InterHuman and Inter-X come with captions, and we captioned HOI-M$^3$ and Panoptic with Qwen3.8-27B from their videos, using the two-stage prompts of SocialGen \citep{yu2026socialgen}.
We do not use the official test splits of InterHuman and Inter-X for training.

\paragraph{In-the-wild video.}
Motion capture covers few settings and records no faces, so the planner also learns from in-the-wild video, and the renderer's adapters learn from it alone.
Our source is the high-quality subset of OpenHumanVid \citep{li2024openhumanvid}, of which 345,642 clips last at least five seconds.
For the renderer, we keep the top quartile by the dataset's aesthetic score, 86,410 clips, take the first 81 frames of each at 16\,fps, and label every frame:
people are detected and tracked with YOLO11 \citep{jocher2024yolo11}, their 2D keypoints are estimated with RTMPose \citep{jiang2023rtmpose}, gaze with the video gaze model of \citet{vuillecard2025enhancing}, and valence and arousal with HSEmotion \citep{savchenko2022classifying} on frontal faces.
For the planner, we need 3D motion, so we scan the clips with a keypoint detector for windows in which people are continuously visible and moving, segment and track each person with SAM~3 \citep{carion2025sam3}, and recover each person's 3D body with SAM 3D Body \citep{sam3dbody2026}.
This yields 120,109 five-second windows from 97,830 clips, with gaze and emotion labeled by the same estimators.
In-the-wild clips keep their original captions.
The planner's training mix is 55.9\% lifted in-the-wild windows, 22.2\% HOI-M$^3$, 11.6\% Inter-X, 7.6\% InterHuman, and 2.8\% Panoptic.

\section{Plan Tokens and Planner}
\label{sec:supp_tokens}

\paragraph{Motion.}
Before encoding, each skeleton is rescaled to canonical bone lengths, and the person's own proportions are written as one body-shape code (256 values) in front of their motion.
The movement of the hips is encoded as a planar velocity, a turning rate, and a height.
Each of the four motion codebooks (three residual levels for the body and one for the head) has 512 entries, and every four frames become four codes, so a five-second clip becomes 21 chunks and 84 motion tokens per person.
The tokenizer is trained on motion capture only and reconstructs validation windows with a root-relative joint error of 38.6\,mm on InterHuman and 25.3\,mm on Inter-X.

\paragraph{Placement and gaze.}
Every other person's position relative to the first person is written on a 3\,cm grid and the heading in 128 bins, at the first frame and again every 16 frames.
Gaze codes carry a validity flag depending on the visibility of the eyes and take one of 80 values.
The count in the header takes values from one to eight, and the two timing tokens take 37 and 29 values.

\paragraph{Planner.}
The plan vocabulary adds 3,748 tokens to the language model.
The tasks are sampled with probabilities 0.5 (caption to plan), 0.2 (plan to caption), and 0.1 for each of three kinds of completion: continuing a plan in time, writing the remaining people given the first ones, and filling masked spans.
The caption is dropped for 10\% of samples to enable classifier-free guidance.
At inference, the timing tokens ask for a plan that starts at zero in a five-second clip, the length of our videos, and the placement tokens are decoded greedily.

\section{Design Choices}
\label{sec:supp_design}

\paragraph{3D or 2D plans.}
Projected to the image plane, the 3D planner matches or exceeds a planner of image-plane keypoints trained with the same recipe, measured with 2D motion evaluators (Inter-X R-precision .608 \vs{} .548, and level on InterHuman), and videos rendered from the two at the same framing have the same visual quality (VBench aesthetic quality .543 \vs{} .535, imaging quality .715 \vs{} .710).

\paragraph{Vision-language or text-only base model.}
Starting from the text-only Qwen3-4B \citep{yang2025qwen3}, which has the same language model architecture, instead of Qwen3-VL-4B gives a slightly worse planner under the same recipe: after 8,000 steps its validation loss on motion tokens is higher, and its R-precision is .520 against .537 on Inter-X and .277 against .286 on InterHuman. This supports our hypothesis that VLMs have better spatial priors than pure LLMs.

\paragraph{Planner or director for emotion.}
\label{sec:supp_affect}
The plan's emotion could be added as a third modality to be written by the planner's motion and gaze generator, trained on the in-the-wild labels, or by a larger language model that reads the caption and planned motion.
To compare these options, we gathered two-person captions that imply an emotion without naming it and annotated the expected expression of each person in each third of the clip.
The generator, trained to also write emotion, meets .59 of these expectations, against .82 for the director. Fine-tuning it on $\sim$8K captions labeled by the director raises it only to .66.
Videos rendered with that generator's emotion are also no better than videos rendered without emotion (.596 \vs{} .592 on the implied expressions and .703 \vs{} .757 on the wrong ones, on the one \affectsmall seed rendered for both).
We therefore adopt the emotion director, which chooses from a list of valence and arousal pairs, and segments are joined with 0.4\,s ramps.

\paragraph{Number of candidates.}
The critic's gain arrives early (\cref{tab:supp_ncand}).
Choosing among four plans already matches or passes the strongest two-person generators (top-1 R-precision .322 on InterHuman and .627 on Inter-X), and choosing among eight raises it from .311 to .335 on InterHuman and from .599 to .641 on Inter-X and lowers FID on both.
More candidates add less, and the 56 we use reach .352 and .661 here (.343 and .658 averaged over 20 generations, \cref{tab:motion}). We adopt 56 for our main results because planning is extremely cheap, but much of the gain can be captured earlier.

\begin{table}[ht]
\centering\small
\setlength{\tabcolsep}{4pt}
\caption{\textbf{Number of candidates.} Motion of the critic's choice among subsets of the same 56 sampled plans, scored as in \cref{tab:motion} but with a single seed per caption. Best in bold.
More than half of the gain arrives by eight candidates.}
\begin{tabular}{@{}ccccc@{}}
\toprule
 & \multicolumn{2}{c}{InterHuman} & \multicolumn{2}{c}{Inter-X} \\
\cmidrule(lr){2-3}\cmidrule(l){4-5}
Candidates & Top-1 $\uparrow$ & FID $\downarrow$ & Top-1 $\uparrow$ & FID $\downarrow$ \\
\midrule
1 & .311 & 0.99 & .599 & 0.42 \\
2 & .315 & 0.87 & .619 & 0.31 \\
4 & .322 & 1.00 & .627 & 0.32 \\
8 & .335 & 0.82 & .641 & 0.24 \\
16 & .347 & 0.86 & .656 & 0.23 \\
32 & .340 & 0.78 & .659 & \textbf{0.20} \\
56 & \textbf{.352} & \textbf{0.73} & \textbf{.661} & 0.22 \\
\bottomrule
\end{tabular}
\label{tab:supp_ncand}
\end{table}

\section{Training Details}
\label{sec:supp_impl}

\paragraph{Planner.}
The planner is trained for 24,000 steps with an effective batch size of 128, learning rate $10^{-4}$ ($2\cdot10^{-4}$ for the embeddings of the new tokens), a 300-step warm-up, and no decay, on four H200 GPUs.
Each training window is cropped at a random offset and mirrored with probability 0.5, with left and right swapped in the caption.
On half of the samples, a random fraction of the input motion codes is replaced with random codes, as in T2M-GPT \citep{zhang2023t2m}.

\begin{table*}[t]
\centering\small
\setlength{\tabcolsep}{4pt}
\caption{\textbf{Questions and answers for one caption}, ``one takes four quick steps forward and embraces the other person'' (\cref{fig:teaser}). A video's VQA accuracy is the fraction of questions answered as expected. The backbone's video fails the count, the action, and the order.}
\begin{tabular}{@{}lp{0.50\textwidth}ccc@{}}
\toprule
Kind & Question & Expected & \method + critic & Wan2.1-T2V-14B \\
\midrule
Action & Does one person take four quick steps forward? & yes & yes & no \\
Interaction & Does the person who takes the steps wrap their arms around the other person in an embrace? & yes & yes & yes \\
Order & Does the person who takes four quick steps forward embrace the other person only after completing those steps? & yes & yes & no \\
Expression & Do both people look warm, affectionate or caring? & yes & yes & yes \\
Wrong expr. & Does either person look angry or hostile? & no & no & no \\
Count & Are there exactly $K=2$ people in the video for (nearly) its whole length, with nobody appearing, vanishing or merging? & yes & yes & no \\
\midrule
\multicolumn{3}{@{}l}{VQA accuracy} & 1.00 & .50 \\
\bottomrule
\end{tabular}

\label{tab:supp_eval}
\end{table*}

\paragraph{Renderer.}
The frozen renderer has 17.3B parameters including the frozen VACE control adapter.
We add eight masked cross-attention layers, one after each of the eight (of 40) DiT blocks that receive VACE's control signal.
A video token inside a face may also attend to one learned null token.
The output projection of each layer starts at zero, so training starts from the unmodified model.
A gaze token embeds a unit 3D direction in camera coordinates, and an emotion token embeds a valence and an arousal, each with a validity flag, and tokens are built for at most four people per clip.
Each face region is the box around the person's nose, eye, and ear keypoints enlarged by 30\% on each side.
The gaze of a person seen from behind is marked invalid.
Each skeleton is drawn on black at $832\times480$ with no reference image.

The adapters were trained for 10,354 steps with batch size 16 and learning rate $7\cdot10^{-5}$ on four B200 GPUs.
They see both gaze and emotion on 68\% of samples, only gaze on 20\%, only emotion on 10\%, and neither on 2\%, with each dropped stream replaced by a learned null token, and the caption is dropped on half of the samples.
The flow-matching loss is weighted four times higher on face pixels and, on gaze-only samples, a further ten times higher on the eyes (3$\times$3 latent cells per eye), with the weights renormalized to mean 1.
Training timesteps are drawn from a logit-normal distribution with mean 2 \citep{esser2024scaling}, and we sample with the matching shift $e^2\approx7.39$ and 30 UniPC steps \citep{zhao2023unipc}.

\section{Plan Critic}
\label{sec:supp_critic}

Candidates are ranked by the sum of their ranks under three critics, each trained with the loss of \cref{sec:critic}. They are four-layer transformers (width 384) over the people's joint positions and velocities in which every motion layer cross-attends to layer 20 of a frozen Qwen3-1.7B \citep{yang2025qwen3}, so that the model can see who does what and in which order.
The first critic's softmax per caption is over the caption's real motion, the planner's own samples for that caption, and the real motions of other captions.
The second critic sees only negatives made of real motion: the real motion of the most similar captions, partial versions of the clip (frozen, or replaced by another clip's motion, from a random point on, or its first part played twice), and staging errors (the partner's motion taken from another clip, the partner turned away or moved apart, the two bodies pushed into each other).
The third sees all negatives: these, six of the planner's samples, and four corruptions of the real motion.
Real motion is first passed through our tokenizer, so that the model cannot separate real from sampled motion by the tokenizer's reconstruction error.

Each model is averaged over three training seeds.
Before ranking, candidates whose trunks pass through each other by more than 8\,cm in a tenth of the frames are excluded.
The three models, the rank sum, and the interpenetration filter were selected on the validation split against single models and other combinations.
On validation, they raise the top-1 R-precision of the motion evaluators from .599 to .660 on Inter-X and from .291 to .320 on InterHuman.

\section{Evaluation Details}
\label{sec:supp_eval}

\paragraph{VQA accuracy.}
For each caption, we use GPT-6.1 Sol to write one to three questions about the expressions the caption implies and about those it rules out (\emph{wrong expressions}). For the 164 captions in which an event implies a shift of gaze, such as looking down before stepping on a foot, the judge also asks about each shift. The judge answers them from 21 frames at 4\,fps and enlarged face crops \citep{zhang2016joint}, blind to which system made the video (example in \cref{tab:supp_eval}).
For \affectsmall, we wrote the questions with the captions, before any plan or video existed. They ask which expression each person shows, when it appears or changes, and which expressions would be wrong.
\emph{VQA accuracy} is the fraction of a video's questions on which the judge's answer matches the caption, averaged over captions.
On \testset, we average two judgments, which agree on 95.7\% of the questions.

\paragraph{Baselines.} Our main baseline is our method's backbone, Wan2.1-T2V-14B. It is the same model without the control branch and the adapters, so the comparison shows what these and the plan add. All models receive the same prompt and seed for each caption.

\paragraph{Camera and motion.}
Our \testset videos use a fixed full shot that keeps every person in frame.
We find that subject consistency, background consistency, and motion smoothness are within .03 of the backbone's. Fewer of our videos pass the motion threshold of VBench's dynamic degree (.31, .35 with the critic, \vs{} .63).
However, this is due largely to the fixed camera. With a camera that follows the people, the critic's plans for the captions reach .94, while aesthetic and imaging quality, subject and background consistency, and VQA accuracy all stay within .01.

\paragraph{Example.}
\cref{tab:supp_eval} lists the questions for the caption of \cref{fig:teaser} and the judge's answers for both videos.
The count question is asked for every caption, and the others depend on what the caption states.
Expression questions are added from the expressions the caption implies.
The judge also writes a short description of each video before answering. For the backbone's video, it notes that ``a third person in blue enters from the right, and all three move into an embrace,'' after which ``the man then appears to merge into the others, leaving two visible people hugging.''

\section{Qualitative Results}
\label{sec:supp_qual}

\begin{figure*}[p]
\centering
\includegraphics[width=\textwidth]{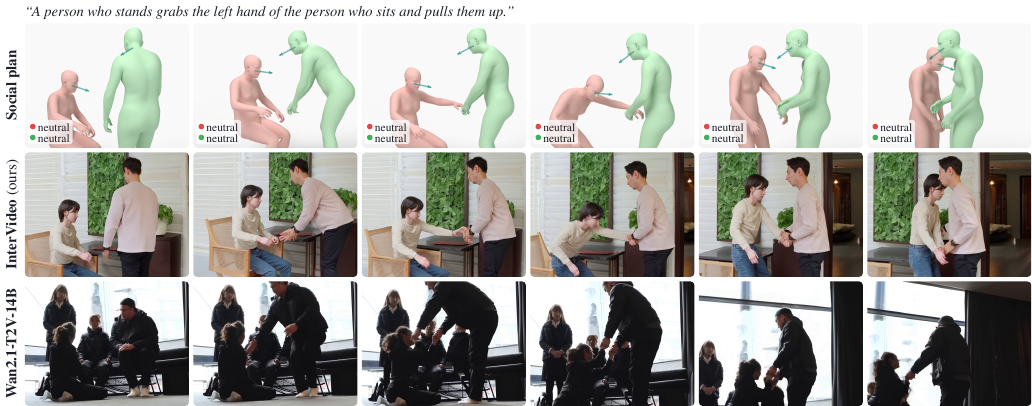}\\[1.2mm]
\includegraphics[width=\textwidth]{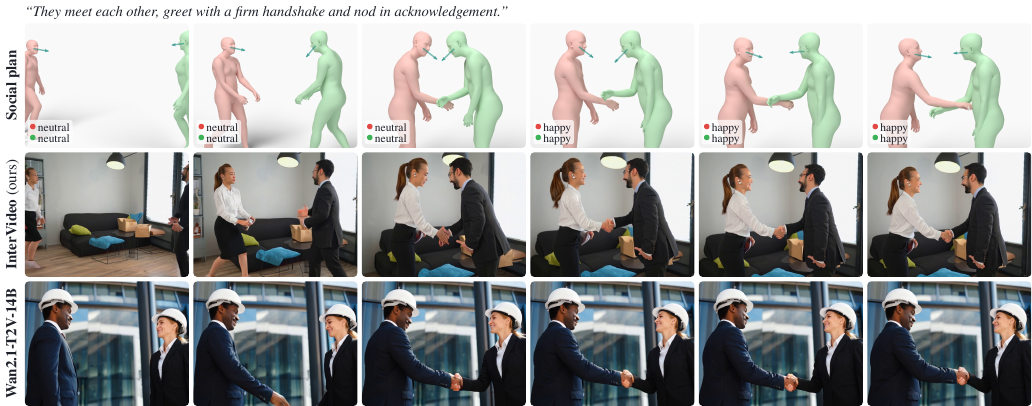}\\[1.2mm]
\includegraphics[width=\textwidth]{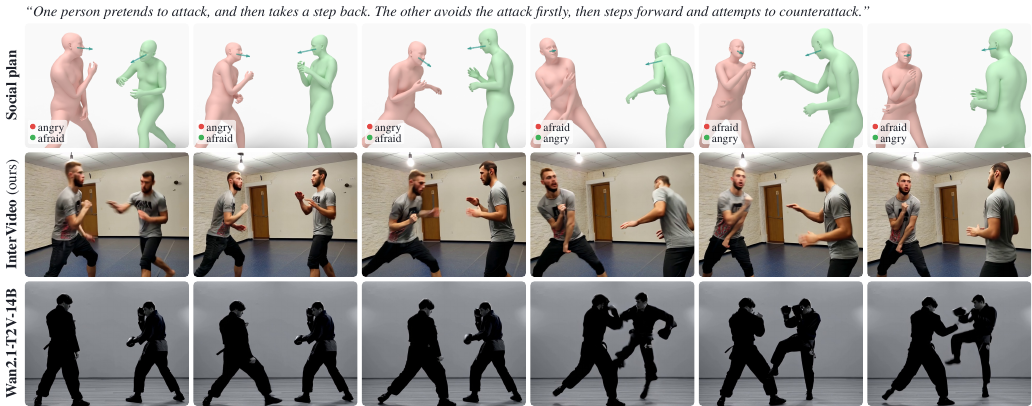}
\caption{\textbf{More qualitative results} given the same captions and seeds, with rows as in \cref{fig:qual} and frames at 0.5, 1.3, 2.1, 2.9, 3.7, and 4.5\,s.}
\label{fig:supp_qual}
\end{figure*}

\begin{figure*}
\centering
\includegraphics[width=\textwidth]{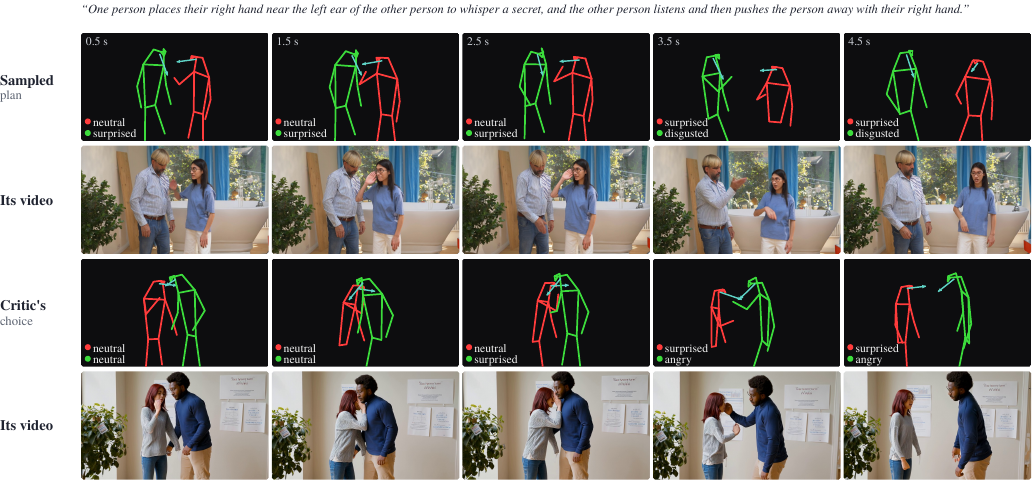}
\caption{\textbf{What the critic changes.} For one \testset caption, a single sampled plan has one person raise a hand beside their own head, so nobody whispers and nobody is pushed.
In the critic's choice among 56, one person leans in to whisper and the other then pushes them away, as the caption asks.
Each video follows its plan.}
\label{fig:supp_critic}
\end{figure*}

\begin{figure*}
\centering
\includegraphics[width=\textwidth]{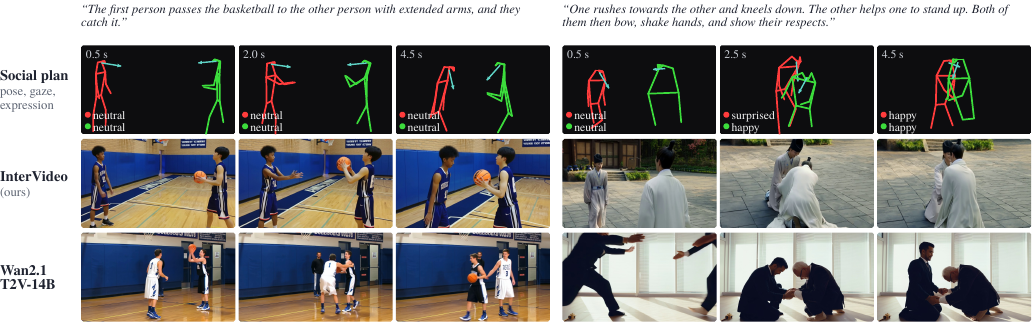}
\caption{\textbf{Failure cases.} Left: the plan has the first person extend their arms, but in our video the other person holds the basketball from the start, and it is never passed. Since the plan does not model objects, human-object interaction is left entirely for the renderer to infer, which points to an important direction for future work.
Right: the caption asks for five events, and our video ends with both people still kneeling.}
\label{fig:supp_fail}
\end{figure*}

\cref{fig:supp_qual} shows more qualitative results, \cref{fig:supp_critic} an example of what the critic changes for one caption, and \cref{fig:supp_fail} two captions on which \method fails.

